# Estimating Blink Probability for Highlight Detection in Figure Skating Videos

Tamami Nakano[†]
Graduate School of Frontiers
Bioscience, Osaka University

Atsuya Sakata
Graduate School of Information
Science and Technology,
Osaka University

Akihiro Kishimoto
Graduate School of Information
Science and Technology,
Osaka University

## ABSTRACT

Highlight detection in sports videos has a broad viewership and huge commercial potential. It is thus imperative to detect highlight scenes more suitably for human interest with high temporal accuracy. Since people instinctively suppress blinks during attention-grabbing events and synchronously generate blinks at attention break points in videos, the instantaneous blink rate can be utilized as a highly accurate temporal indicator of human interest. Therefore, in this study, we propose a novel, automatic highlight detection method based on the blink rate. The method trains a one-dimensional convolution network (1D-CNN) to assess blink rates at each video frame from the spatio-temporal pose features of figure skating videos. Experiments show that the method successfully estimates the blink rate in 94% of the video clips and predicts the temporal change in the blink rate around a jump event with high accuracy. Moreover, the method detects not only the representative athletic action, but also the distinctive artistic expression of figure skating performance as key frames. This suggests that the blink-rate-based supervised learning approach enables high-accuracy highlight detection that more closely matches human sensibility.

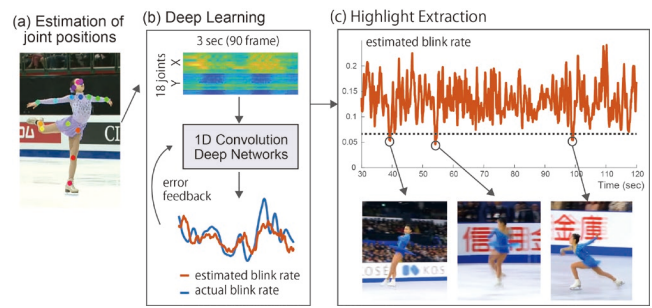

Figure 1: An overview of the proposed highlight detection system

## 1 Introduction

With the vast number of videos being uploaded to the Internet, the need for highlight detection and video summarization to save time is increasing yearly. However, searching for and editing important scenes from a video is time-consuming and costly. Therefore, the automatic extraction of highlight scenes has been extensively investigated [1]. The unsupervised learning approach based on inherent physical features in videos dominates in highlight detection methods [2-9]. However, this approach is limited by the possibility that the extracted highlight may not match people's interests. Hence, supervised approaches based on human-related information such as manual evaluation, facial expression, vocalization, body movements and physiological reactions have also been developed [10-14], but subjective human evaluation is costly and unreliable [15]. Additionally, highlight detection requires a high time accuracy, but emotion and arousal related physiological indices change relatively slowly in units of seconds. This necessitates an index that has a higher temporal accuracy and objectively reflects human attention and interest. Therefore, we focus on spontaneous human blinking behaviors. The previous study found that spontaneous blinks were synchronized within 0.2 s among people viewing the same video [16, 17]. The people suppressed their need to blink during the important scenes, while they simultaneously blinked at the attentional breaks in the video. Moreover, the higher the interest in the video, the higher the blink synchronization among people [18]. Therefore, viewers' blink rate can be used as an objective and temporally accurate indicator of the scenes they are most interested in. In this study, we propose a new method for automatic highlight detection based on the blink probability rate estimated by artificial neural networks.

Highlight detection has been studied especially in sports videos because of its wide viewership and high commercial potential [19-24]. Particularly, the success of the detection of highlights can be analyzed using figure skating performances, in which characteristic action events such as jumps and spins are embedded in continuous movements for several minutes [21, 25]. We therefore used figure skating performance videos as the target of highlight detection. First, we created the datasets of figure skating competition videos and measured the blink rate of people as they watched the videos. Next, the skater's joint positions were identified (Figure 1(a)), and their spatio-temporal changes were input to a three-layer one-dimensional convolutional neural network (1D-CNN) to estimate the blink probability rate (Figure 1(b)). After learning, the CNNs estimated the blink rate for each frame of the novel video, and identified the scene where the blink rate was lower than the threshold as the highlight (Figure 1(c)).

The three principal contributions of this study revolve around automatic highlight detection in sports videos. First, we introduce people's blink rate to highlight detection as an objective index of the level of human interest, and this is unprecedented. Second, blink rate enables the detection of highlight scenes with high time accuracy even with a simple neural network structure. Finally, the proposed method matches human sensibility better than the previous methods because it detects not only the representative athletic action but also the distinctive artistic expression of the figure skating performance. The remainder of this paper describes these contributions in detail.

## 2 Related Works

Extensive studies have developed methods for extracting moments of special interest from videos. The existing methods of highlight detection can be classified into three approaches. First, the most common method is an unsupervised approach of analyzing the video's inherent spatio-temporal characteristics. Additionally, to reflect human interest and emotion on highlight detection, a supervised approach using human-related information has also been researched. Finally, there have been recent methods of artificial neural networks for automatically detecting highlights. In this section, we briefly review related works on highlight detection from these three representative approaches.

### 2.1 Video features-based highlight detection

The unsupervised learning approach of analyzing information inherent in video dominates the field of video summarization and highlight detection [1]. Most of these methods analyze spatiotemporal features of videos such as motion energy, saliency, object recognition, repetition, and loudness of sound [5, 7, 8]. The representative images were also identified by clustering-based methods [2, 6, 9], and the dictionary learning methods [3, 4]. Additionally, a method based on content-specific video structure analysis such as play/break detection in soccer games was found effective for highlight detection especially in sport videos [19, 20, 22]. These methods are advantageous for being able to detect highlights only with video, but they have the disadvantage of being based solely on physical features that do not necessarily reflect human interest.

### 2.2 Human reaction-based highlight detection

In the supervised learning approach for highlight detection, the standard method involves a higher ranking video segment based on manual human evaluations, which is assigned as the highlight [10, 11, 26]. However, manual human evaluations are time-consuming, costly, and unreliable [15]. Quantitative human-related information, which includes facial expressions, heart rate, and body movements, can be used as an effective index for highlight detection [12-14, 27]. However, since these behavioral indices change depending on factors, such as arousal, emotion, fatigue, etc., time accuracy is not high in these indices, and it is unclear whether they sharply reflect the interest level of the audiences.

### 2.3 Artificial neural networks for highlight detection

Artificial neural networks are recently used for highlight detection in both unsupervised and supervised learning approaches. The unsupervised learning approach extracts images with few reconstruction errors as key frames [28, 29], while the supervised approach identifies video highlights by learning interest scores using Recurrent Neural Network or LSTM [30, 31]. Furthermore, the artificial neural networks for an evaluation of action performance can also be applied to highlight detection by identifying the images which impact the quality score of an action [21, 25]. Although these complex artificial neural networks have improved the accuracy of identifying characteristic action scenes, they do not guarantee that they can capture the viewer's complex attention mechanisms.

## 3 The Proposed Method

To extract highlights from figure skating performance videos, this study proposes a simple 1D-CNN that learns the correspondence between the blink probability rate and the spatio-temporal feature of the skaters' actions. The blink rate, which is the percentage of people blinking at each video frame, was measured in advance before the CNN learning. In order for CNN to efficiently capture the characteristics of the skater's movements, we used the temporal change of the skater's joint position as input data. The skater's 18 joint positions were estimated using OpenPose [32] for each video frame, and the X and Y coordinate positions across 3 seconds (90 frames) were converted into a 90 × 36 image and input to the CNN. As shown in Figure 2, the neural network consisted of three convolution layers followed by the average pooling layer and fully connected layer. The kernel size of all convolutional layers was set to 8 and the number of filters was set to 64-128-64. The CNN estimated the blink probability rate at the last frame of the input (the 90th frame). Optimization learning was conducted by error feedback between the actual and estimated blink rates. After the convergence of the learning, the blink probability rate for the new figure skating video was estimated, and the time points when the value was lower than the threshold was detected as the highlight scenes (Figure 1(c)).

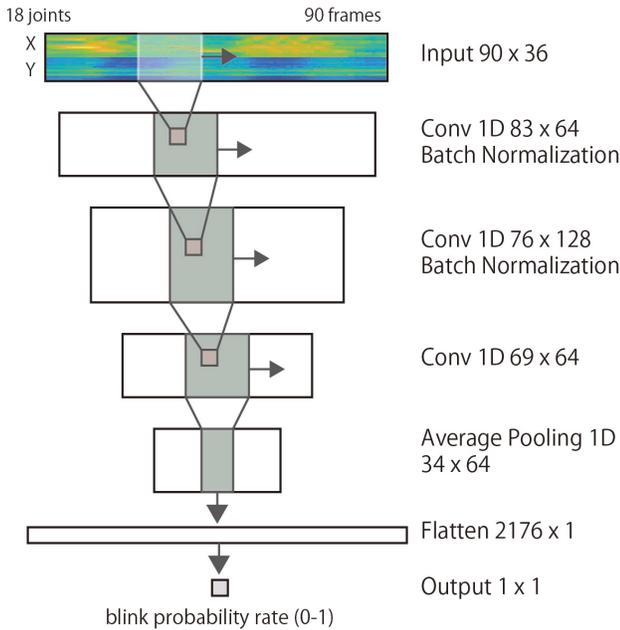

Figure 2. Architecture of the 1D-CNN

## 4 Experiments

### 4.1 Dataset

Forty-eight video clips of figure skating performance by a single skater were used in this study. The duration varied between the video clips (mean 3 min 46 s, min-max 2 min 40 s – 4 min 48 s,30 FPS), and the total time is about 180 min. These are competition videos for broadcast in the Olympics or formal international figure skating matches. Thirty-three video clips are performances of female skaters and 15 video clips are performances of male skaters. Event times of jump and spin were manually listed in each video to verify whether the network model was able to extract these events correctly later.

To extract the spatio-temporal features of the skater's action movement, their joint positions for each video frame were identified using OpenPose toolbox [32]. One limitation was that OpenPose detected the joint positions of audiences in addition to the skaters on the video images. To address this problem, taking advantage of the skaters' usual focus in the videos, the joint information of the person whose body area was the largest in each video frame was selected. Additionally, the joint data with a low confidence score of estimation (< 0.7) was excluded, and interpolated it with the data before and after. The time series data of the X and Y positions of 18 joints were extracted for 3 s (90 video frames) with a sliding window of 1 s. A total of 112,039 samples were created for the input data of neural networks.

People's blinking behaviors while watching the video clips were analyzed for use as an indicator of audience interest. Their pupil diameters were monitored using the near infrared light eye tracker (Tobii Pro Spectrum, Tobii AB, Sweden) at a sampling frequency of 120 Hz. The blink onset time for each participant was extracted by detecting a combination of rapid increase in pupil diameter followed by a rapid decrease within 0.5 s [16]. After that, the percentage of people blinking was analyzed in each video frame. The blink rate of 24 video clips was calculated from a data sample of 38 participants, and other 24 video clips were calculated from a data sample of 12 participants. All participants were college students in our university. The blink rate in the last frame of the joint time series dataset (36 joint XY positions x 90 video frames) was adopted as the blink rate for that dataset.

Because the size of the dataset was not large, the leave-one-out cross-validation method was applied for learning in the neural networks. The dataset extracted from 47 out of 48 video clips was used as training data, and the remaining one video clip was used as test data. The blink rate in each frame of the test video clip was estimated by the neural networks.

### 4.2 Network architecture and implementation

Our system was implemented in Python on a workstation with an Intel ® Xeon CPU® 3.0 GHz 4-cores CPU and NVidia Tesla P100 graphic cards. The detailed neutral network architecture of the CNN is described in Section 3. The estimation of blink rate was optimized by minimizing the root mean squared error (RMSE) between the ground truth and predicted blink rate. The network was optimized by Adam optimizer with learning rate of 0.001 [33]. We set the maximum epoch number to 100 and a batch size of 4,096 samples.

### 4.3 Performance evaluation

We examined the correlation between the time series data of the predicted blink rate by neural network and those of actual blink rate in each video clip. In our dataset, even the shortest video clip has 4,800 observation points (2 min 40 s). When there are many observation points in the time series data, the RMSE becomes very small, and there is a high possibility that a significant pseudo-correlation is detected even if there is no correlation. To address this problem, we created the surrogate blink rate by randomly shuffling the time-series of the predicted blink rate. This randomized data preserved the exact blink rate distribution but lost the temporal structure. We repeatedly calculated the correlation coefficient between the surrogate estimated blink rate and the actual blink rate, and created a distribution of 1000 random correlation coefficients. Using the mean and variance of this distribution, we tested a significance of the actual correlation coefficient using t-test (multiple comparisons were corrected using Bonferroni method; alpha level = 0.05/48). Consequently, 45 of the 48 video clips (94%) had a significant positive correlation between the estimated blink rate and the actual blink rate (Figure 3, red circles).

Next, we focused on the skating jump event which grabs the attention of audiences and induces a characteristic temporal change in the blink rate. As shown in the blue line of Figure 4(a), in the case of two continuous jump events, the rapid decrease in the blink rate further continues two times in association with each jump. The blink rate estimated by the neural networks is also able to correctly estimate two rapid declines in these two consecutive jumps (orange line in Figure 4(a)). We compared the temporal change of the blink rate between the estimated and the actual data from 1 s before to 3 s after all the 272 jump events. As shown in figure 4(b), both the estimated and the actual blink rate started to decrease gradually from 0.5 s just before the jump, showed a sharp decline with the jump and increased 1 s after the jump. We conducted the shuffle-corrected statistical correlation test for the 272 time-series data. The estimated blink rate was significantly positively correlated with the actual blink rate in 72 % of the jump events. These results demonstrate that the present neural networks almost succeeded in estimating the temporal change in the blink rate from the spatiotemporal features of the skater's actions.

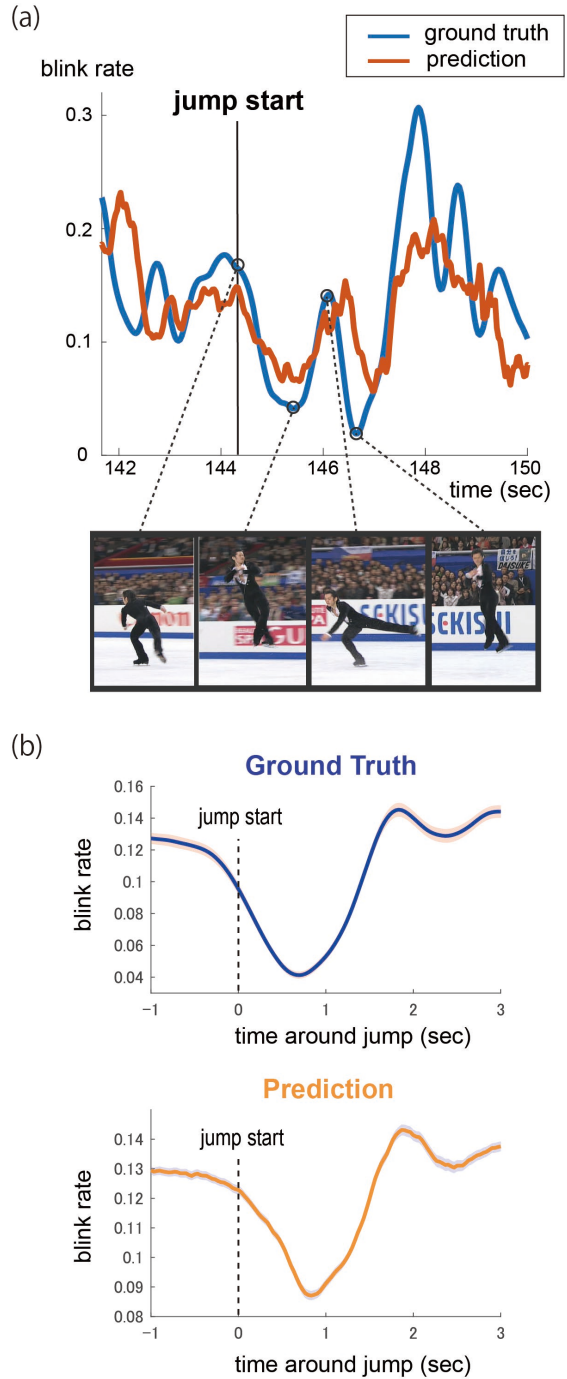

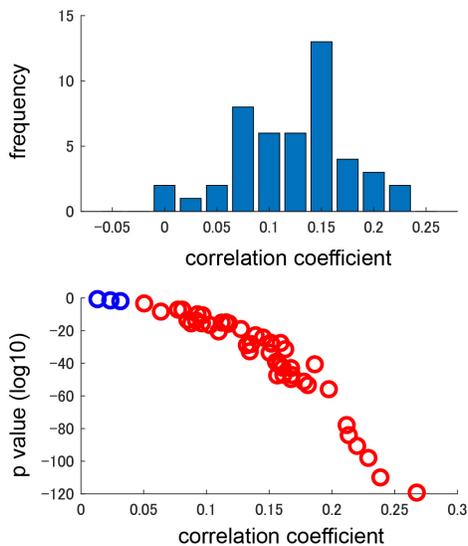

Figure 3. Distribution of correlation coefficients and their p-values

Figure 4. Actual and estimated blink rate around jump events. (a) a representative example (b) the average blink rate across all jump events. Shaded area represents a standard error from the mean.

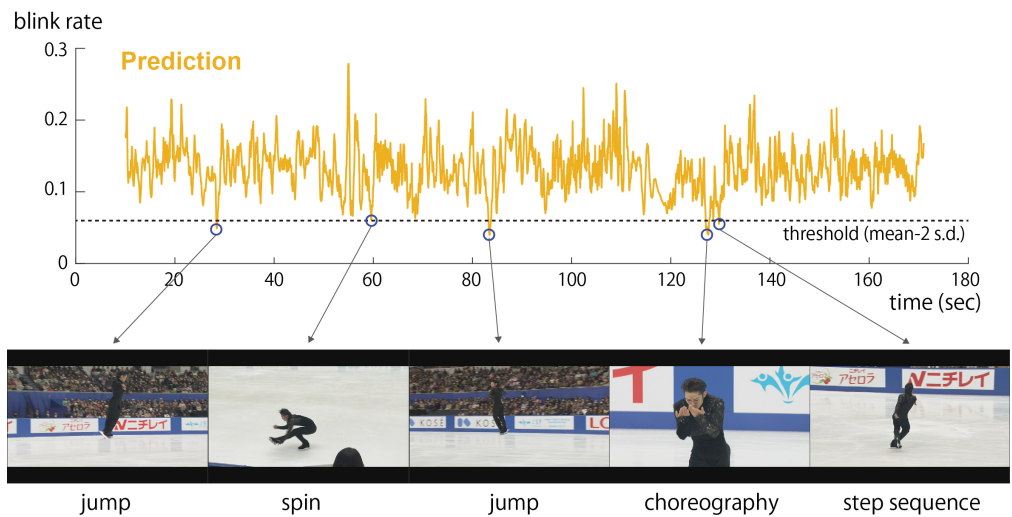

Figure 5. Highlight detection based on the predicted blink probability rate

## 4.4 Highlight detection

People's blinks are suppressed at the important scenes of videos and occurs simultaneously at the attentional breakpoints [16, 17]. This suggests that a point at which the blink probability is significantly low is expected to be an important scene in the video. We therefore defined highlight scene where the estimated blink probability is at least two standard deviations below the mean across 5 video frames (Figure 5). Thus, 340 scenes were detected, which were mainly related to important actions in figure skating competitions such as jumps and spins (7.1 ± 3.2 s.d. scenes on average were detected from each performance video). Further, to evaluate which category these detected scenes belonged to, we first created the short video clips covering from one second before to one second after the detected scenes. Then, two naïve evaluators watched these short video clips and categorized the actions of the skater into five categories (jump, spin, step sequence, distinctive choreography, just sliding). Their ratings agreed at a rate of 90%. As shown in Figure 6, jump accounted for nearly half (48%) of the detected scenes. It was followed by distinctive choreography, which accounted for 26%. Spin and step sequence accounted for 9% each. The remaining 8% were scenes where nothing happened (just sliding). Further, 66% of the detected scenes were related to the critical figure skating actions for technical evaluation scores and 26% were related to artistic expressions. Figure skating performance was evaluated based on both athletic skill and artistic expression. The previous methods for highlight detections have successfully detected action events such as jumps and spins, but have failed to detect highlights related to artistic expression [21, 25]. In contrast, the present method has succeeded in detecting the highlight scenes related to both factors. This is because the blink probability rate directly reflected the interest level of people who watched the figure skating performances with interest in both athletic and artistic factors.

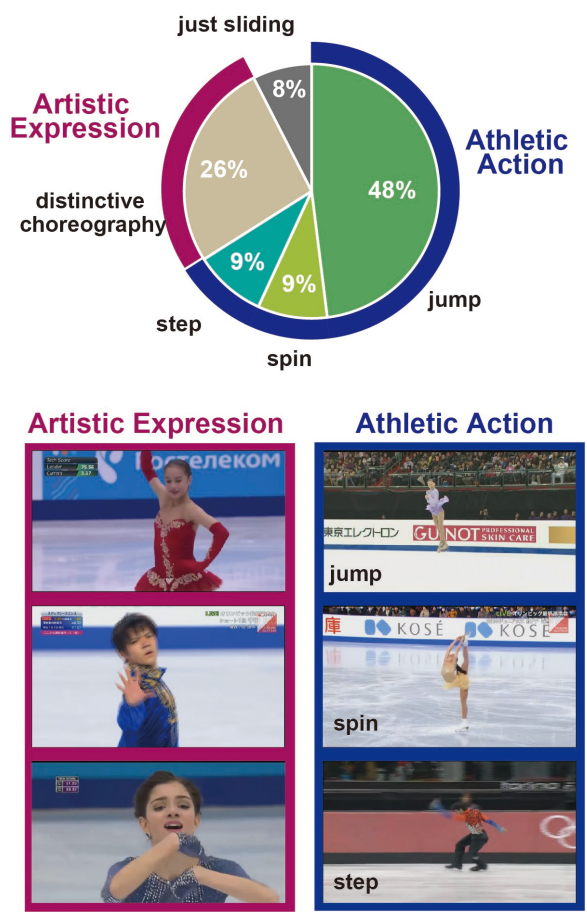

Figure 6. Percentage of the category of highlights (upper) and examples of highlights belonging to artistic expression and athletic action (bottom)

## 4.5 Discussion

*4.5.1 Comparison to other human-related information.* This study succeeded in detecting momentary highlight events in figure skating performances, such as jumps that ended in less than one second, by utilizing information from peoples' blinking behaviors. Previous studies have used physiological activities such as heart rate and respiratory activity as human-related indicators of highlight detection [12, 27]. However, since there is a delay of a few seconds before physiological activity reflects changes in autonomic nervous system activity, it is quite difficult to detect highlights with a high temporal accuracy of less than one second. Human physical activities such as facial expression, vocalization, and body movement are also used as an indicator of highlight detection [5, 13]. However, these physical reactions often occur after the end of a highlight event (e.g., after a soccer goal was scored). For this reason, it is difficult to specify the start time and end time of an event with high temporal accuracy. Additionally, people often watch videos still and silent even when they have a strong interest in it. These physical indexes can capture large changes in excitement and arousal, but have difficulty capturing subtle temporal changes in human interest level. In contrast, spontaneous blinks are controlled directly by the central nervous system according to visual attention [34], so the blink rate changes sharply in response to momentary highlight events. In addition, since people always spontaneously generate blinks, we can stably capture a temporal change in the human-interest level under any state of arousal. Furthermore, blinking behavior can be easily monitored from outside using a web camera [18]. Based on the above points, the instantaneous blink rate is superior to other human-related information as an index of the viewer interest level in terms of temporal accuracy, easy measurement, and high robustness.

*4.5.2 Comparison to previous methods.* Another important result of this study is that the artificial neural networks that has learned the relation between spatio-temporal features of video and blink rate can detect not only the athletically distinctive events but also the artistically distinctive events. The attractiveness of figure skating lies not only in its acrobatic movements, but also in its elegant choreography. However, while the previous studies using performance scores of figure skating by professional judges as teaching signal has been successful in detecting athletic highlights such as jumps and spins, it has failed to extract highlights related to artistic elements [21, 25]. Because the timing of blinking is controlled by attention based on not only human cognition but also emotion, the blink rate sharply reflects viewer's emotional reaction to the skater's performance. In these regards, this study demonstrates that development of artificial neural network systems that learn human blinking behaviors is very effective in extracting highlight scenes that match human sensitivity.

## 5 Conclusion

In this paper, we present a novel automatic highlight detection framework that utilizes blink rate as an objective indicator of human interest. By extracting the spatio-temporal changes of the skater's joint positions in advance, we succeeded in estimating the blink probability rate of figure skating videos with a simple CNN structure and high temporal accuracy. The major improvement that this method achieves over the conventional method is that the present method can detect not only the events of athletic actions such as jumps, but also the events of choreography that are artistically noticeable. This suggests that the supervised learning approach based on human blinking behaviors can extract highlight scenes that fit both human attention and sensibility. The blinking behavior can be easily measured with a webcam, thus creating a large data set does not require significant cost and effort. In the future, we need to verify that this method can be applied to other sports videos and daily life videos.

## ACKNOWLEDGMENTS
This work was supported by a Grant-in-Aid 18H04084 awarded to TN from the Ministry of Education, Culture, Sports, Science and Technology, Japan, as well as a PRESTO grant, "Design of information infrastructure technologies harmonized with societies" #11027, awarded to TN from the Japan Science and Technology Agency (JST), Japan.